\documentclass{article}

 \usepackage[preprint]{neurips_2025}
 \usepackage{natbib}

\usepackage[utf8]{inputenc} 
\usepackage[T1]{fontenc}    
\usepackage{hyperref}       
\usepackage{url}            
\usepackage{booktabs}       
\usepackage{amsfonts}       
\usepackage{nicefrac}       
\usepackage{microtype}      
\usepackage{xcolor}         
\usepackage{amsmath}
\usepackage{booktabs}
\usepackage{caption}
\usepackage{graphicx}
\title{No-Knowledge Alarms for Misaligned LLMs-as-Judges}

\author{%
  Andr\'es Corrada-Emmanuel\thanks{andres.corrada@dataengines.com} \\
  Data Engines \\
  \texttt{andres.corrada@dataengines.com} \\
}

\begin{document}

\maketitle

\begin{abstract}
  If we use LLMs as judges to evaluate the complex decisions of other LLMs,
  who or what monitors the judges? Infinite monitoring chains are inevitable
  whenever we do not know the ground truth of the decisions by experts and
  we do not want to trust them. One way to ameliorate our evaluation
  uncertainty is to exploit the use of logical consistency between
  disagreeing experts. By observing how LLM judges agree and disagree
  while grading other LLMs, we can compute the only possible evaluations
  of their grading ability. For example, if two LLM judges disagree on
  which tasks a third one completed correctly, they cannot both be
  100\% correct in their judgments. This logic can be formalized
  as a Linear Programming problem in the space of integer response
  counts for any finite test. We use it here to develop no-knowledge
  alarms for misaligned LLM judges. The alarms can detect, with no
  false positives, that at least one member or more of an ensemble
  of judges are violating a user specified grading ability requirement.
\end{abstract}

\section{Introduction}

No intelligence, however powerful, can escape the laws of logic and physics 
(\cite{tegmark2023}). We can use logical consistency between disagreeing
LLM judges to ameliorate the problem of infinite monitoring chains whenever
ground truth about the correctness of LLM decisions is absent. This can
occur as an economic choice - we choose to automate a process with LLMs
and want to further automate supervision by using other LLMs as judges.
Or it can occur as an ignorance problem on our part - we are trying to
supervise LLMs smarter than us or ones doing tasks we cannot understand.
Either case is a no-knowledge situation where logical consistency, not
correctness, is the only tool available for formalizing evaluations.

We can build a logic of unsupervised evaluation for the task of grading
multiple choice exams. The central question in any logic of unsupervised
evaluation is -- what group evaluations are logically consistent with
how we observe experts agreeing and disagreeing in their decisions? This
computation can be formalized for multiple-choice test takers or,
equivalently, classifiers. It is easy to be seduced into thinking that
a logic formalism for classification tests has nothing to do with
LLMs carrying out complex tasks and evaluations. The problem of
infinite monitoring chains, however, makes multiple choice exams
inevitable in any holistic evaluation framework that, seemingly
paradoxically, should include the evaluators themselves.

As LLMs evolve in capability, evaluations have to evolve with them.
For this reason, the development and validation of suitable benchmarks
and evaluation frameworks is an active area of research. Given the range
of tasks they can perform, holistic evaluation frameworks, like HELM 
(\cite{liang2023}) measure multiple metrics across different 
"dimensions" of performance (e.g. accuracy, calibration, robustness, 
fairness, bias, toxicity, and efficiency).

The problem of evaluating LLMs at scale has an economic side and
an epistemological one. The principal-agent problem (\cite{principalagentWiki})
in economics is concerned with incentives when a \emph{principal} delegates
a task to an \emph{agent} that may have conflicting interests and objectives.
The epistemological problem with evaluating LLMs occurs whenever we want
to scale it across number of evaluations or ability of the LLMs.

We can deal with a large number of evaluations by delegating the evaluation
task to an agent -- the idea behind the concept of LLM-as-a-Judge (\cite{chang2023}). 
But this approach cannot resolve the problem of evaluation at scale whenever the
principal is unable to judge the agents because the ground truth for the correctness
of their decisions is unavailable. This occurs when we are trying to evaluate
agents more intelligent than us and we do not understand their work. But it also
occurs everyday in the industrial use of LLMs when we do not have the time or
money to use panels of human judges to judge their quality in production 
(\cite{chang2023}).

This paper proposes to ameliorate the epistemological issues of the principal-agent
monitoring problem by using logical consistency between disagreeing LLMs. If two LLMs as judges
disagree on their grades for a third worker LLM, for certain evaluation domains, we can take this
as conclusive proof that they cannot both be 100\% correct in their grading. This simple
example is a \emph{no-knowledge} alarm. Having observed the one-bit of information that
the LLM judges disagreed in their grades, with no knowledge of the test or its correct answers,
we can correctly deduce that one or both did not grade perfectly. Before discussing
the construction of the proposed logic, it is important to note the limits of what
any logic can do in unsupervised settings.

Logical consistency cannot detect that \emph{agreeing} LLMs are correct. In the
simple example above, if we had observed the two LLM judges agreeing on their
evaluations of a third, we would not be able to logically exclude both of them being
perfect, as unlikely as that may be. That
can only be done when we have logical correctness -- in supervised settings where the
ground truth for the test answers is available. In that sense, 
logical consistency between members of an
ensemble is acting like error-correcting codes where only some bit errors can be detected.

Logical consistency is also unable to establish the validity of any evaluation.
Validation of LLM evaluations and the use as LLM-as-a-Judge remains a scientific 
and engineering problem that is actively being pursued 
(e.g.\ \cite{salaudeen2025, Szymanski2025}). The rest of the paper
will discuss how to generalize the one-bit alarm to multiple-choice tests under
the assumption that they are valid and have a ground truth, however imperfectly
that is defined. 

\section{A no-knowledge alarm based on logical consistency}

The use of the alarm, not its validity or usefulness, will be
demonstrated by considering 25 pair comparisons from the MT-Bench
benchmark (\cite{chang2023}). Technical details on how this
set was selected can be found in section \ref{sec:appendix} but we
will outline the rationale for the selection here.

Unsupervised evaluation is going to occur where, by definition,
we do not have the answer key for the correct decisions of
LLM-as-judges grading the complex output of assistant LLMs
as is done in the MT-Bench dataset. The complexity of those
assistant LLM outputs means that human experts do not agree
on which is best from a pair or if both are equally good.
\cite{chang2023} reports 80\% agreement on these three grading
labels.

But say we wish to confirm that LLM judges are aligned with
the ground truth established by some majority voting between
human experts. There are about 5K human judgments for pair comparisons
of LLMs in the MT-Bench human judgments accessible via HuggingFace.
Restricting ourselves to those that contain 2 or more human
votes, we are left with 269 pair comparisons. When
we further restrict that set with the decisions of two graders, gpt4-pair
(denoted by \texttt{gpt4}) 
and the majority judgments of the authors of Chang et al\. (\texttt{authors}), we
get our evaluation set of 25 pair comparisons. The confusion matrices
for the grades given out by \texttt{gpt4} and \texttt{authors} are shown in Tables 1
and 2.

The small numbers of this toy example are the result of using
the MT-Bench benchmark to establish a fixed set of pair comparisons
for two graders (\texttt{gpt4} and \text{authors}) for which we have a
somewhat reasonable ground truth (pair comparisons where we can
weigh the votes of two or more human experts). The example, however,
is sufficient for illustrating how the logic of the alarm works.

\subsection{Points in the  \texorpdfstring{$Q$}{Q}-simplex for a classification test}

We are going to use the unknown counts of each label in the ground
truth for a test of size $Q$ to carry out the algebraic logic
of the alarm. This maps our ignorance of the answer key for
a classification test onto an integer space of dimension equal
to the number of labels in our classification test. A point in
this space is an integer tuple of size $R$, the number of possible classification
responses.

Every classification test that we give, lies somewhere in this
space that we will call the $Q$-simplex since the integer
counts of labels in the answer key must obey the equality,
\begin{equation}
    \label{eq:q-simplex}
    \sum_{\ell_{\text{true}} \in \mathcal{L}} Q_{\ell_{\text{true}}} = Q.
\end{equation}
Statements about logical consistency between disagreeing experts can
only be made at fixed points on this simplex. Even though we
do not know where we are on the $Q$-simplex for any given test
that we carry out, we can reason given the assumption of being
at a fixed point in the $Q$-simplex. If using that reasoning,
we detect that at every possible point in the $Q$-simplex for
a test, we can deduce the same thing, that thing must be true
for the test.

In our running example, we simulate an unsupervised evaluation
setting by considering the observable counts of the question
aligned decisions by two graders -- gpt4 and authors -- to
show how their grading disagreements can alert us that one
or both are violating the condition that they must be
at least 50\% or better in picking the average human expert answer.

\begin{table}[h!]
\centering

\begin{minipage}{0.48\textwidth}
\centering
\caption{Grading confusion matrix for \texttt{gpt4}}
\begin{tabular}{lccc}
\toprule
 & \multicolumn{3}{c}{\textbf{Decision Labels}} \\
\cmidrule(lr){2-4}
\textbf{True grades} & model a & model b & tie \\
\midrule
model a & 1 & 0 & 3 \\
model b & 2 & 9 & 3 \\
tie & 2 & 1 & 4 \\
\bottomrule
\end{tabular}
\end{minipage}
\hfill
\begin{minipage}{0.48\textwidth}
\centering
\caption{Grading confusion matrix for \texttt{authors}}
\begin{tabular}{lccc}
\toprule
 & \multicolumn{3}{c}{\textbf{Decision Labels}} \\
\cmidrule(lr){2-4}
\textbf{True grades} & model a & model b & tie \\
\midrule
model a & 2 & 1 & 1 \\
model b & 1 & 12 & 1 \\
tie & 1 & 5 & 1 \\
\bottomrule
\end{tabular}
\end{minipage}

\end{table}

Table~1 shows \texttt{authors} did badly grading pair comparisons where human experts
preferred model a (25\% correct), better on model b grading (64\% correct), and
marginally better on tied pair comparisons (57\%).
The grades of gpt4-as-a-judge (Table~2) were, 
in comparison,  50\% aligned on model a
pair comparisons, 86\% correct on model b pair comparisons, and 14\% on tied
comparisons. We can summarize this by saying that both graders are misaligned
on at least one label at the $Q$-simplex for the ground truth
$Q$-simplex point, $(Q_a=4,Q_b=14,Q_t=7)$.

In an unsupervised setting, all we would have for these two graders would
be their pair comparison grades. Given that three grades are
possible in these comparison of assistant LLM outputs, we could see
9 possible ways these two graders could agree or disagree on any given
pair comparison.
The observed count of these 9 possible pair grades for our running example
are shown in Table~3. Since 3 of the possible vote patterns are not observed
we will continue development of the alarm using the votes of each grader.
These are $(Q_a=5,Q_b=10,Q_t=10)$ for \texttt{gpt4} and $(Q_a=4,Q_b=18,Q_t=3)$
for \texttt{authors}. We can view these are the decision by the graders of
what is the ground truth $Q$-simplex point for the pair comparisons they
just graded ($(Q_a=4,Q_b=14,Q_t=7)$).

\begin{table}[h!]
\centering
\caption{Counts for agreements and disagreements of two graders
(\texttt{gpt4} and \texttt{authors}) on 25 pair comparisons from the MT-Bench dataset}
\begin{tabular}{ccccccccc}
\toprule
(a, a) & (a, b) & (a, t) & (b, a) & (b, b) & (b, t) & (t, a) & (t, b) & (t, t) \\
\midrule
3 & 2 & 0 & 0 & 10 & 0 & 1 & 6 & 3 \\
\bottomrule
\end{tabular}
\end{table}

\subsection{The set of correct grade counts at a fixed point in the  \texorpdfstring{$Q$}{Q}-simplex}

Even though we do not know where on the $Q$-simplex an actual test is, we can reason
given any assumed fixed point on the $Q$-simplex. In particular, we can determine
the set of possible number of correct grades for each grader given that assumption.
This can be done in two ways. The first way is to invoke linear constraints that 
require the number of correct
to be between 0 and some maximum count determined by our observations. For example,
since \texttt{gpt4} only has 5 model a grades, it never could be more correct than
that. In general, for grader $i$, we can impose the constraints defined by
\begin{equation}
    0 \geq R_{\ell_i, \ell_{\text{true}}} \geq R_{\ell_i}.
\end{equation}

But, in addition, for three label classification tests we have three linear equations
that must be obeyed by all possible number of correct grades for each classifier. 
For label a, this linear equation is,
\begin{equation}
    \label{eq:a-axiom}
    R_{a_i,a} = Q_a - R_{b_i} - R_{t_i} + R_{b_i,b} + R_{b_i,t} + R_{t_i,t} + R_{t_i,b}.
\end{equation}
The two other label equations are obtained by permuting the labels in this equation.

\subsection{The logic of the alarm}

Assuming a fixed point in the $Q$-simplex, we can use the constraints and linear equations
to see if we can exclude number of correct grades by the two graders where both are better
than 50\% accurate on all grade labels. This is easily demonstrated at extreme points like
on the $Q$-simplex like $(Q_a=25,Q_b=0,Q_t=0).$ At this point, \texttt{gpt4} can only be
$5/25$ correct -- we have detected that \texttt{gpt4} is misaligned at this $Q$-simplex point.

We can scan through all the points in the simplex and ask the same question -- are
both graders better than 50\% correct? If the answer is no at all points on the $Q$-simplex,
we have detected that one or more of the graders are misaligned with the unknown answer key.
This is the case for our running example. We illustrate the set of possible number of correct
grades in Figure~1 for the $Q$-simplex point, $(Q_a=8,Q_b=9,Q_t=8).$ Note that the graders
are misaligned for labels a and t at this assumed point in the $Q$-simplex.

As the fixed point in the $Q$-simplex is changed, which grader is misaligned can change.
Therefore, this alarm cannot tell us which of the graders is the one that is misaligned.
It just knows that both cannot be simultaneously aligned with any possible unknown answer key.

\begin{figure}[h!]
\centering
\includegraphics[width=\textwidth]{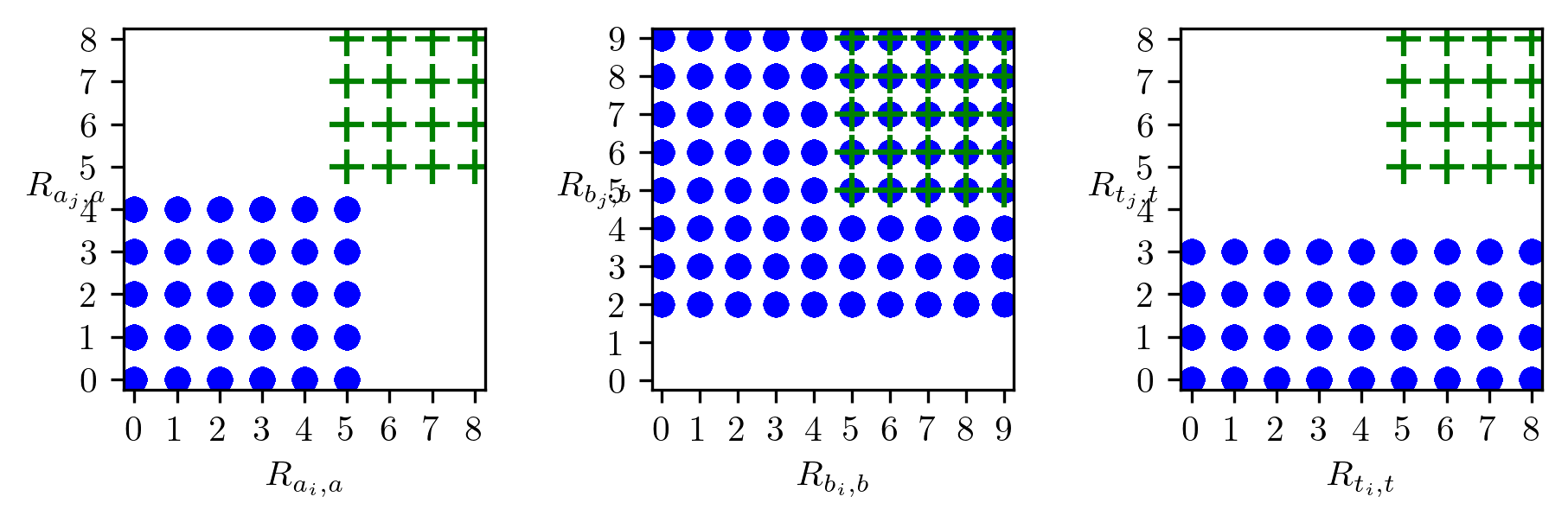} 
\caption{Possible number of correct grades by \texttt{gpt4} and \texttt{authors}
given the assumption that the answer key for the 25 MT-Bench pair comparisons 
contains $(Q_a=8,Q_b=9,Q_t=8)$ labels.}
\label{fig:example}
\end{figure}

\section{Conclusion}
A running example using 25 pair comparisons from the MT-Bench dataset were used to illustrate
how the counts of grades by graders can differ enough that we can logically prove one or both
must be less than 50\% accurate picking the correct grade. This demonstration cannot
establish the safety and/or validity of using these alarms to construct safer AI systems.

\section*{References}
\medskip
{
\small
\bibliographystyle{plainnat}
\bibliography{references}

}


\appendix

\section{Technical Appendices and Supplementary Material}
\label{sec:appendix}

\subsection{Related work}
The logic presented here has a precursor in the \emph{agreement equations} of Platanios
(\cite{Platanios:2014}). The agreement equations were derived using logic alone as
is done here for our label axioms. The axioms presented here, however, go beyond
the agreement equations. To understand if one has identified a complete set of axioms
for unsupervised evaluation it is necessary to work on the projective space of
average performance, not the integer space we have been using in this paper.

The completeness of axioms of unsupervised evaluation comes from the fact
that given $N$ test takers we have a finite and complete set of possible
agreements and disagreements between them ($R^N$) when there are $R$ possible
responses. For each of these possibly observable agreement/disagreement events
we can write a polynomial expression relating it to the unknown statistics
of prevalence and accuracy on each labels. These expressions can represent
the outputs of any ensemble, even highly correlated ones, by including
unknown statistics for the possible correlations between them.

This paper focused on the $m=1$ axioms - the universal set of linear equations
in response space for a single test taker. These axioms where enough to detect
misalignment in the paper's running example. But as noted earlier, if we observed
that the experts answered so their summary statistics were equal, no misalignment
would be detected. But if we include the $m=2$ axioms, we could discover that
they were misaligned.

A simple example shows how this works. Suppose we had observed both LLM judges
grading such that both had summary statistics of their grades that were identical.
Say, they both said $(Q_a=10, Q_b=0, Q_t=10).$ The m=1 axioms would not detect
misalignment with this case. Using the $m=2$ axioms, however, involves looking
at question-level alignment of the two LLM judges. Using those statistics of
pair grades, we could detect that the judging pair is misaligned. For example,
we may find that every time judge 1 said ``model\_a'' the other said ``model\_b''
and so on so we observe that $R_{a_1,a_2} = 0$, $R_{b_1,b_2} =0$, and
$R_{t_1,t_2}=0.$ The LLM judges never agree in their grades, in other words.
This would trigger an alarm that used the $m=2$ axioms.

\subsection{Error-correlated LLM judges}
The issue of correlated LLM judges is not a problem with a logic that seeks to
find all logical consistent evaluations for them. Error-correlated group evaluations
are included in the possible set, along with error-independent ones. The
determination of these possible error-correlations comes in a ladder-like manner.
Using the $m=1$ axioms in this paper, we determine all points in the response
simplex corresponding to single LLM judge responses.

To find the possible error-correlations between two LLM judges we then need to
invoke the $m=2$ axioms to compute all possible pair responses given true label.
These are terms of the form,
\begin{equation}
    R_{\ell_1, \ell_2, \ell_{\text{true}}}.
\end{equation}
The error correlations between LLM judges can then be computed for any
point in the set of possible LLM judge responses. Take the case of
error-correlation on ``model\_a'', that would be computed as,
\begin{equation}
    \frac{R_{a_1,a_2,a}}{Q_a} - \frac{R_{a_1,a}}{Q_a} \frac{R_{a_2,a}}{Q_a}.
\end{equation}

\subsection{Construction of the 25 MT-Bench pair-comparisons grading ground truth}

The majority vote of human experts in the MT-Bench pair comparisons was done by
weighted voting. An expert vote was split into two $1/2$ votes in the case
of ties in a pair comparison, but kept as $1$ if model a or model b was picked.
Thus, for example, if two experts voted (model a, tie) this would be consider
a preference for model a. Since we are using the tie grading label, this procedure
is unambiguous in assigning one of the three available grades.
The same procedure was used to establish the grades by the panel of \texttt{authors}.
In that case, we allowed pair comparisons that had just grades by one author. The
turn 1 pair comparisons used are show in Table~4.

\begin{table}[htbp]
\label{human-ground-truth}
\centering
\caption{Human expert weighted vote ground truth for the 25 LLM pair
comparisons from the MT-Bench dataset. There are 4 model\_a grades, 14
model\_b, and 7 ties.}
\begin{tabular}{c c c c}
\toprule
Question id & Model A & Model B & Grade \\
\midrule
82  & gpt-4            & claude-v1        & model\_b \\
82  & llama-13b        & gpt-4            & model\_b \\
85  & vicuna-13b-v1.2  & gpt-3.5-turbo    & tie \\
91  & gpt-3.5-turbo    & gpt-4            & tie \\
91  & llama-13b        & gpt-3.5-turbo    & model\_b \\
91  & vicuna-13b-v1.2  & gpt-3.5-turbo    & model\_a \\
98  & vicuna-13b-v1.2  & gpt-3.5-turbo    & tie \\
99  & vicuna-13b-v1.2  & gpt-3.5-turbo    & model\_b \\
103 & gpt-3.5-turbo    & gpt-4            & model\_b \\
104 & alpaca-13b       & gpt-3.5-turbo    & tie \\
105 & gpt-3.5-turbo    & gpt-4            & model\_b \\
111 & llama-13b        & gpt-4            & model\_b \\
112 & gpt-3.5-turbo    & claude-v1        & model\_a \\
115 & gpt-3.5-turbo    & claude-v1        & model\_a \\
121 & vicuna-13b-v1.2  & gpt-4            & model\_b \\
124 & gpt-3.5-turbo    & gpt-4            & model\_b \\
134 & gpt-3.5-turbo    & gpt-4            & model\_b \\
140 & alpaca-13b       & gpt-3.5-turbo    & tie \\
141 & gpt-3.5-turbo    & gpt-4            & tie \\
142 & gpt-3.5-turbo    & gpt-4            & model\_b \\
143 & alpaca-13b       & vicuna-13b-v1.2  & tie \\
148 & gpt-3.5-turbo    & gpt-4            & model\_b \\
148 & llama-13b        & claude-v1        & model\_b \\
151 & llama-13b        & gpt-3.5-turbo    & model\_b \\
152 & alpaca-13b       & vicuna-13b-v1.2  & model\_a \\
\bottomrule
\end{tabular}
\end{table}

\subsection{Suspecting the ground truth}

Holistic evaluation frameworks already contain dimensions of evaluation
for which there is no easy way to establish the ground truth as
shown in the somewhat arbitrary construction of it discussed in
the previous section. This is not just an artifact of the small size
of the set used here ($Q=25$). \cite{chang2023} report an
overall agreement of 80\% between human grades in their 5K+
dataset of human pair comparison judgments.

Logical necessity is not magical, it does not have access to
some unknown ground truth. We should include in any misalignment
detection the possibility that there is no ground truth, in fact,
for the test the experts took. Or maybe we should suspect the
ground truth as, for example, offered here.

The misalignment detection algorithm can be viewed as saying that
there is no answer key for the present test that also has
the experts fulfilling our grading ability requirements. So if
there is a ground truth for the test, we would have to conclude
the experts are misaligned.

But, the test may be the problem or any purported answer key
that another expert offers us. In that case, the semantic
free nature of logical consistency between counts can naturally
include any suggested answer key as just one more test taker.
Now suppose two LLM judges agree sufficiently that the misalignment
alarm is not triggered, but when we include our answer key, we
do. Logical necessity cannot establish who is the wrong party
here. It could be the answer key, or it could be the two LLM judges.

And finally, suppose the two LLM judges and the answer key are
aligned enough the alarm is not triggered given our grading
quality specification. Logical necessity cannot warn us in
cases where all experts are aligned on a mistaken ground truth.

\subsection{Proof sketch of the label ``a'' axiom}

The equality in the label ``a'' axiom relating observed responses by a classifier
and its unknown by true label responses can be established by expanding
quantities of the form $R_{\ell_i}$ by 
$\sum_{\ell_{\text{true}}} R_{\ell_i,\ell_{\text{true}}}$ while at the same time
expanding $Q_a$ as
\begin{equation}
    Q_a = R_{a_i, a} + R_{b_i,a} + R_{c_i,a}.
\end{equation}
The axiom then follows trivially by linear cancellation of terms.

\end{document}